\documentclass[11pt,a4paper]{article}
\usepackage[hyperref]{naaclhlt2019}
\usepackage{times}
\usepackage{latexsym}
\usepackage{helvet}  
\usepackage{courier}  
\usepackage{graphicx}  
\usepackage{footmisc}
\usepackage{multibib}
\usepackage{caption}
\usepackage{multirow}
\usepackage{graphicx}
\usepackage{subfigure}
\usepackage{makecell}
\usepackage{bm}
\usepackage{CJKutf8}
\usepackage{array}
\usepackage{booktabs}
\usepackage{url}
\usepackage{color}

\aclfinalcopy 

\title{Recovering Dropped Pronouns in Chinese Conversations via \\ Modeling Their Referents}

\author{Jingxuan Yang,
  Jianzhuo Tong,
  Si Li $^*$,
  Sheng Gao,
  Jun Guo,
  Nianwen Xue $^\dag$\\
  Beijing University of Posts and Telecommunications \\
  $^\dag$Brandeis University \\
  {\tt \{yjx, tongjianzhuo, lisi, gaosheng, guojun\}@bupt.edu.cn} \\
  {\tt xuen@brandeis.edu}}

\date{}

\begin{document}
\begin{CJK*}{UTF8}{gbsn}
\maketitle
\begin{abstract}
Pronouns are often dropped in Chinese sentences, and this happens more frequently in conversational genres as their referents can be easily understood from context.
Recovering dropped pronouns is essential to applications such as Information Extraction where the referents of these dropped pronouns need to be resolved, or Machine Translation when Chinese is the source language.
In this work, we present a novel end-to-end neural network model to recover dropped pronouns in conversational data. 
Our model is based on a structured attention mechanism that 
models the referents of dropped pronouns utilizing both sentence-level and word-level information.
Results on three different conversational genres show that our approach achieves a significant improvement over the current state of the art. \let\thefootnote\relax\footnotetext{$^*$ Corresponding author}
\end{abstract}

\section{Introduction}
Chinese is a pro-drop language, meaning that it is not always necessary to have an overt pronoun when the referent is clear from the context. This is in contrast with a non-pro-drop language like English, where an overt pronoun is always needed. For example, Kim~\shortcite{Kim:00} shows that an overt subject is used only in 64\% of the cases in Chinese while that percentage is over 96\% in English.

Even in pro-drop languages like Chinese, pronouns are omitted to different degrees in different genres. Pronouns are dropped more often in informal conversational genres than in formal genres like newswire. Recovering these dropped pronouns (DPs) is important to applications such as Machine Translation where dropped pronouns need to be made explicit when Chinese is translated into a target language \cite{wang:2016a, wang:2016d, wang:2018} or Information Extraction where relations might involve entities referred to by these DPs. 

\begin{figure}
	\centering
	\includegraphics[width=7.7cm, height=5.5cm]{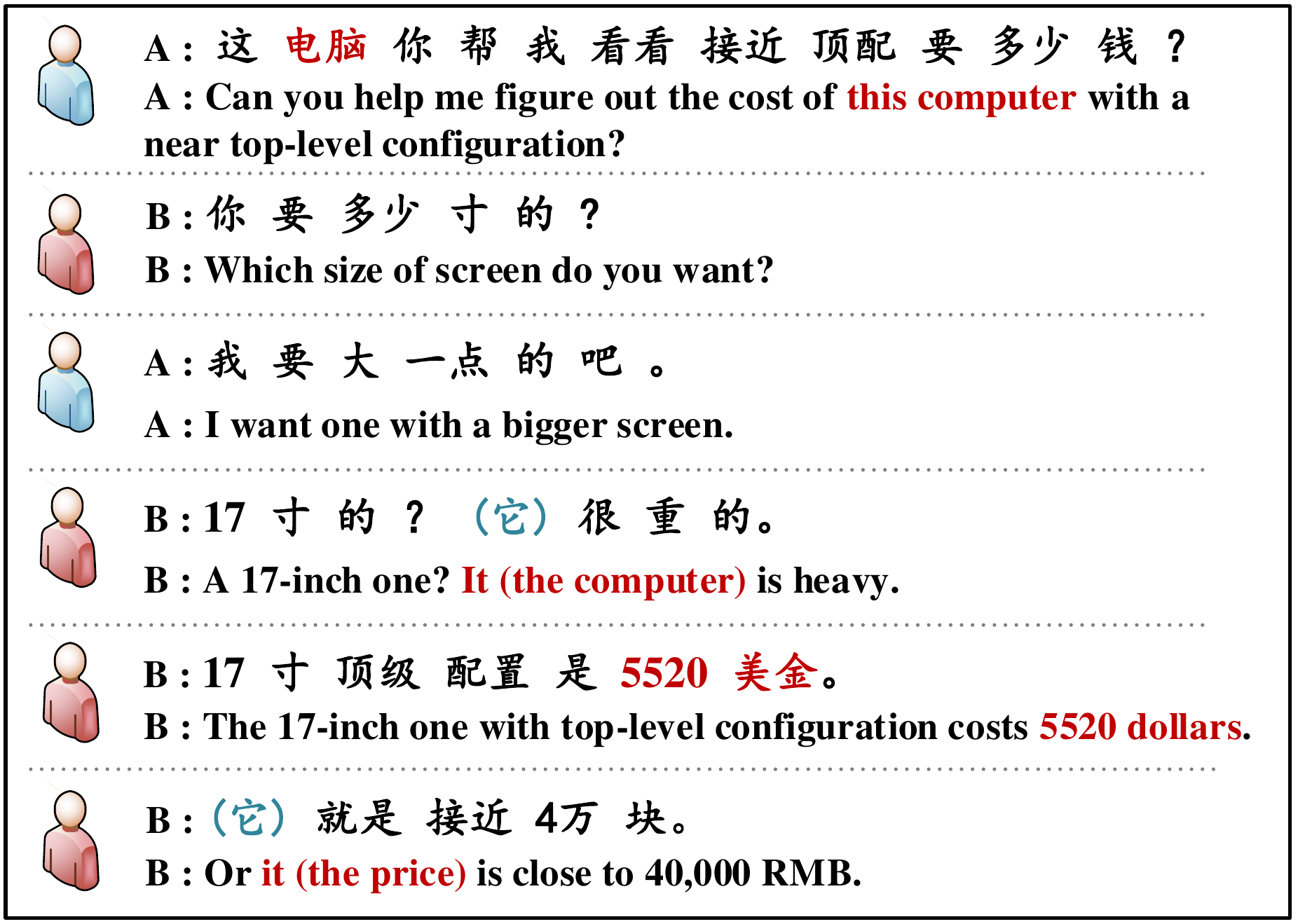}
	\caption{An example conversation between two people. The first dropped pronoun refers to ``this computer" in the first sentence and is translated into ``it" in English. It needs to be recovered from the wider context.}
	\label{conv_example}
\end{figure}

More concretely, recovering DPs involves i) first locating the position in a sentence where a pronoun is dropped, and ii) determining the type of the pronoun that is dropped. This is illustrated in Figure~\ref{conv_example}, where the recovered dropped pronouns are in parenthesis. Both instances of dropped pronouns can be replaced with the overt pronoun~它~(``it''), but they refer to different things in the conversation.
Dropped pronoun recovery is different from zero pronoun (ZP) resolution \cite{Chen:15,Yin:17, yin:2018} where the focus is on resolving an anaphoric pronoun to its antecedent, assuming the position of the zero pronoun is already determined. Here we do not attempt to resolve the dropped pronoun to its referent. Dropped pronoun recovery can thus be viewed as the first step of zero pronoun resolution, but this task is also important in its own right. In applications like machine translation, dropped pronouns only need to be identified and translated  correctly but they do not need to be resolved.

Traditionally dropped pronoun recovery has been formulated as a sequence labeling problem where each word in the sentence receives a tag that indicates whether the word has a dropped pronoun before it and which pronoun is dropped.
Yang et al.~\shortcite{Yang:15} leverages lexical, contextual and syntactic features to detect DPs before each word from a predefined list of pronouns. Giannella et al.~\shortcite{giannella:17} utilizes a linear-chain conditional random field (CRF)  classifier to simultaneously predict the position as well as the person number of a dropped pronoun based on lexical and syntactic information. These featured-based methods require labor-intensive feature engineering. 

In contrast, Zhang et al.~\shortcite{zhang:neural} uses a multi-layer perceptron based neural network model to recover dropped pronouns, which eliminates the need for feature engineering. The input to their model is a context embedding constructed by concatenating the embeddings of word tokens in a fixed-length window. However, the referent of a dropped pronoun that provides crucial information to determine the identity of a pronoun is typically found outside the local context of the dropped pronoun, and thus cannot be effectively modeled with a window-based multi-layer perceptron model. 
For example,  in Figure \ref{conv_example} the referent of the first dropped pronoun, the third person singular pronoun 它~(``it'') 
is 电脑~(``computer''), which appears in the very first utterance, several utterances before the pro-drop sentence. Therefore, long-distance contextual information needs to be captured in order to determine the type of the dropped pronoun.

\begin{figure*}
	\centering
	\includegraphics[width=16cm, height=8cm]{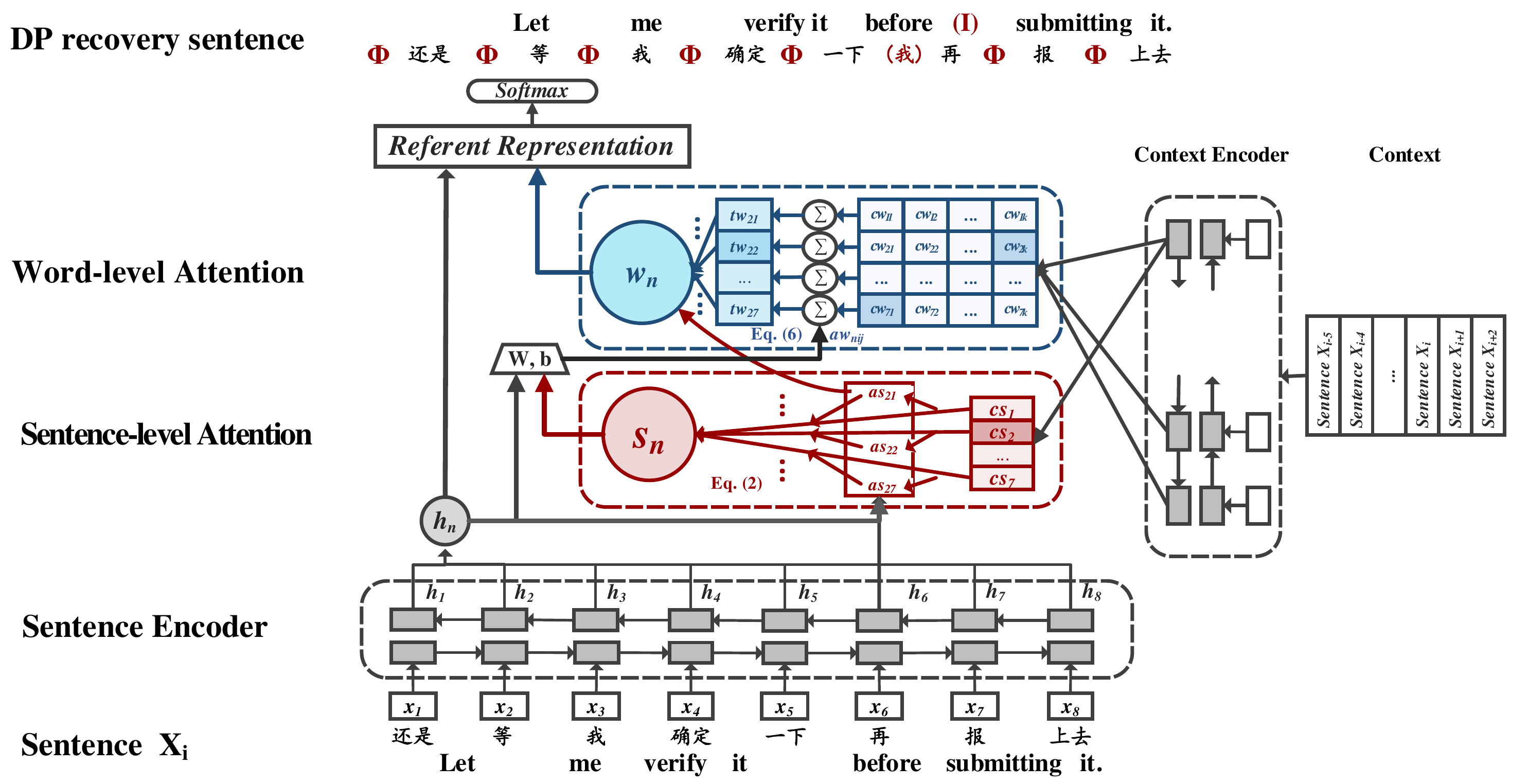}
	\caption{\label{framework} Neural Dropped Pronoun Recovery Framework.}
	\label{overall}
\end{figure*}

In this paper, we describe a novel Neural Dropped Pronoun Recovery framework, named NDPR that can model the referent information in a much larger context. The model makes use of contextual information at two levels of granularity: the sentence level and the word level.
An illustration of the NDPR framework is given in Figure~\ref{framework}. The referent modeling process is implemented with a structured attention mechanism. For each word token in the pro-drop sentence, the model attends to the utterance in which the referent of the dropped pronoun is mentioned by sentence attention 
and then zeros in to the referent by word attention. The resulting referent representation is combined with the representation of the dropped pronoun to make the final prediction on the location and type of the dropped pronoun. We demonstrate the effectiveness of our proposed model on three Chinese conversational datasets and results show that our approach outperforms the current state of the art by a fairly large margin.

We also perform ablation studies to explore the contribution of the different components of our model. We show that word-level attention is more effective in recovering {\it concrete} pronouns because it serves as a matching mechanism that matches the representation of the dropped pronoun against that of the DP referent. In contrast, sentence-level attention can be considered as an auxiliary tool for word-level attention to improve model accuracy by filtering out irrelevant referents. All code is available at \texttt{\url{https://github.com/ningningyang/NDPR}}. 

Our contributions are summarized as follows:
\begin{itemize}
\setlength{\itemsep}{-2pt}%
	\item We propose a novel attention-based neural network to recover dropped pronouns in Chinese conversational data by modeling their referent.
	\item We evaluate our system on three conversational genres and show that our model consistently outperforms the state of the art by a large margin on all three datasets. 
	\item We also present experimental results that demonstrate the effectiveness of various components in our framework and analyze some mistakes made by our model on each dataset.
\end{itemize}

\section{Related Work}
\subsection{Dropped pronoun recovery.} Dropped pronoun detection originates from Empty Category (EC) detection and resolution, a task aimed to recover certain dropped elements in syntactic treebanks \cite{chung:10,cai:11,xue:13}. Dropped pronoun recovery was first proposed as an independent task in \cite{Yang:15}, which leveraged a set of specially designed features to recover DPs in Chinese text messages. Giannella et al.~\shortcite{giannella:17} employed a linear-chain CRF classifier to jointly determine the position and person number of dropped pronouns in Chinese SMS messages using hand-crafted features. These traditional feature-based methods require heavy feature engineering. Zhang et al.~\shortcite{zhang:neural} for the first time utilized a multi-layer perceptron model to recover dropped pronouns. Each word is expressed as a concatenation of embeddings of word tokens in a fixed-length window. This method can not resolve the cases when referents are outside the local context, which is a common occurrence in conversational data. There has also been some work that attempts to recover dropped pronoun in pro-drop languages to help Machine Translation \cite{wang:2016a, wang:2016d, wang:2018}. There are some inherent differences between their work and ours. 
In their work, they attempt to create training data by projecting pronouns from another language to Chinese, and the specific location of a dropped  pronoun may not matter as much as long as it is translated correctly when Chinese is the source language. 
In contrast, we focus on recovering DPs from conversational context, and determining the correct position of a dropped pronoun is critical part of the task.

\subsection{Zero pronoun resolution.}  A line of research that is closely related to our task is zero pronoun (ZP) resolution, which aims to resolve Chinese pronouns to their antecedents. 
Converse and Palmer~\shortcite{Converse:06} presented a rule-based method to resolve ZP by utilizing the Hobbs algorithm \cite{Hobbs:78}. Learning-based anaphoric resolution approaches have also been extensively explored. Zhao and Ng~\shortcite{Zhao:07} and Kong and Zhou~\shortcite{Kong:10} proposed systems that perform ZP resolution by integrating syntactic features and position information in a system based on decision trees and context-sensitive convolution tree kernels. 
With the powerful learning capacity of neural networks, recent work focuses on learning representations to resolve zero pronouns or common noun phrases \cite{Ng:07,Yin:17,yin:2018}. 
The DP recovery task explored in our work focuses on determining the position and type of dropped pronoun without attempting to resolve it to an antecedent.

\subsection{Attention and memory network.} The attention mechanism is first proposed for neural machine translation \cite{Bahdanau:14} and has been attempted in a variety of natural language processing tasks \cite{Bahdanau:14,Hermann:15}. Kim~\shortcite{Kim:16} extended the basic attention mechanism to incorporate structural biases by combining graphical models with neural networks. Yang et al.~\shortcite{Yang:16} proposed a word and sentence-level hierarchical attention network for document classification. Xing et al.~\shortcite{xing2018} also presented a hierarchical recurrent attention network (HRAN) to model the hierarchy of conversation context in order to generate multi-turn responses in chatbots. 
In contrast with these bottom-up mechanisms, our model 
adopts a top-down structured attention mechanism to construct the representation for a DP. Our model attempts to identify utterances which contain the referent of a DP, and then focus in on the representation of the referent itself. 
Our model draws inspiration from the memory network and its variants \cite{Weston:14,Sukhbaatar:15,Henaff:16,Miller:16}, where an external memory component is used to store and update knowledge. 

\section{Approach}
Following \cite{Yang:15}, we formulate DP recovery as a sequential tagging problem. Given an input sentence $X=(x_1,x_2, ... ,x_s)$ and its contextual utterances $\bm{C}=(X_{1},...,X_{m})$, our task is to model $P(Y|X, \bm{C})$ and predict a set of labels $Y=(y_1,y_2, ... ,y_s)$. Each element of $Y$ indicates whether a pronoun is dropped and which type it should be before each word in the sentence $X$. The NDPR model consists of three components: (1) Input Layer; (2) Referent Modeling Layer; (3) Output Layer. We describe them in detail below.

\subsection{Input Layer}
This layer converts the raw inputs $X$ and $\bm{C}$ into distributed representations. 

\subsubsection{Sentence $X$}
During the encoding process, the input sentence of length $s$, is transformed into a sequence of $d$-dimensional word embeddings at first, and then fed into two different kinds of bidirectional recurrent neural networks (RNN) \cite{Elma:91}:
\begin{itemize}
\setlength{\itemsep}{-2pt}%
	\item \textbf{BiGRU} \cite{Bahdanau:14}: Each word $x_n, n \in \{1, ...,s\}$ of the sentence is represented as a concatenation of forward and backward hidden states as: $\bm{h_n} = [\bm{\overleftarrow{h}_n}, \bm{\overrightarrow{h}_n}]$. We aim to express DP by the hidden state of the first word after DP.
	\item \textbf{PC-BiGRU} \cite{Yin:17}: Pronoun-centered BiGRU (PC-BiGRU) also contains two independent GRU networks. For each word $x_n, n \in \{1, ...,s\}$, the forward GRU$_f$ encodes the preceding context of DP from left to right as $\bm{\overrightarrow{h}_{n-1}}$, and the backward GRU$_b$ models the succeeding context as $\bm{\overleftarrow{h}_n}$ in the reverse direction. Final hidden state of DP is also concatenated as $\bm{h_n} = [\bm{\overleftarrow{h}_n}, \bm{\overrightarrow{h}_{n-1}}]$. The intuition of this structure is to express DP by the last word before DP and the first word after DP.
\end{itemize}

\subsubsection{Context $\bm{C}$}
The context provides the necessary background information to recover DPs. In our work, we utilize five utterances preceding the one the DP is in and two utterances following current input sentence $X$ as context $\textbf{C}$. The size of $\textbf{C}$ is determined empirically since our statistics show that $97\%$ of the dropped pronouns can be inferred based on information in the 7 sentences surrounding the sentence in which the dropped pronoun occurs. The context $\textbf{C}$ passes through the same encoder as $X$, yielding two kinds of memories: (1) sentence-level memory: concatenated final states of the forward and backward GRU for each contextual utterance $i=\{1,...,m\}$ as $\bm{cs_i} = [\bm{\overleftarrow{cs}_i}, \bm{\overrightarrow{cs}_i}]$, holding sentence level background knowledge. (2) word-level memory: a set of concatenated hidden states at each time step $j=\{1,...,k\}$ as $\bm{cw_{i,j}} = [\bm{\overleftarrow{cw}_{i,j}}, \bm{\overrightarrow{cw}_{i,j}}]$, expressing the contextual information of words in memory.

\begin{figure*}
	\centering
	\includegraphics[width=16cm, height=5.2cm]{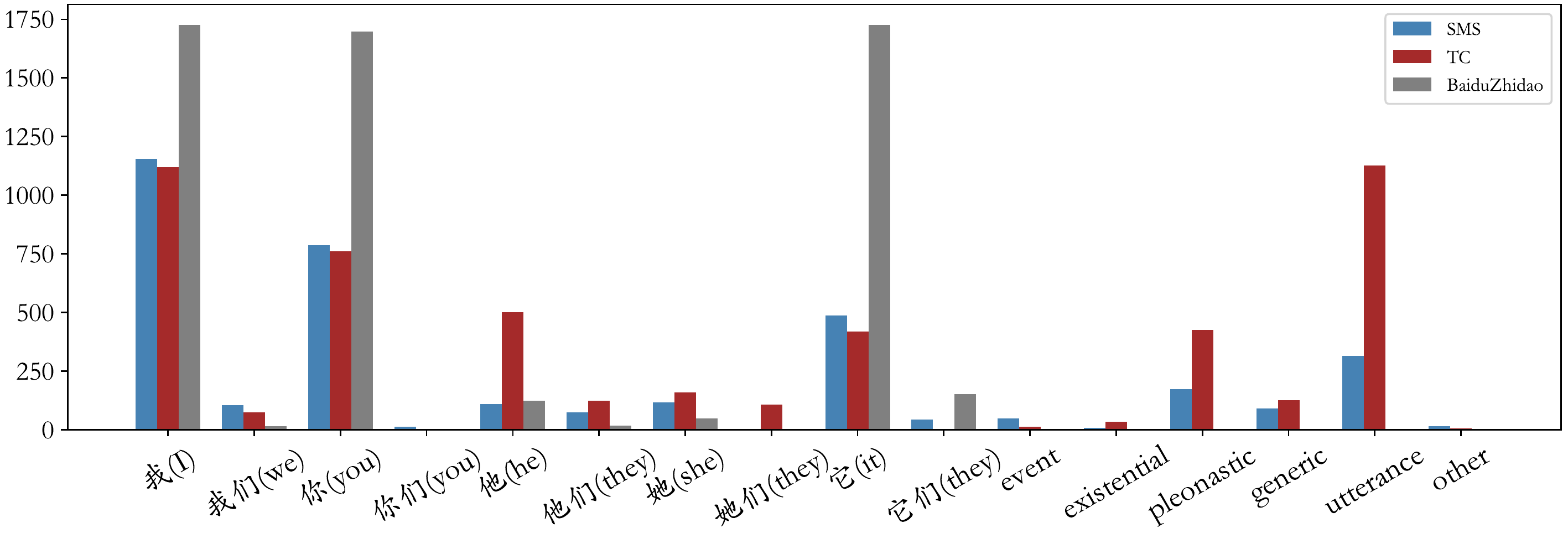}
	\caption{Counts of each type of dropped pronouns in different conversational data sets. Pronouns have a more balanced distribution in Chinese SMS and telephone conversation(TC). For BaiduZhidao, only concrete pronouns are annotated and the distribution of these pronouns is extremely uneven.}
	\label{corpus-statistics}
\end{figure*}

\subsection{Referent Modeling layer}
The referent modeling layer models the interaction between the input sentence $X$ and its context $\bm{C}$ and outputs representation of DP referents. The relevant information for predicting the dropped pronoun is retrieved by a two-step attention-based inference mechanism in which the intermediate results of sentence-level attention is utilized to support subsequent word-level inference. 

\subsubsection{Identify utterances mentioning DP} 
This operation aims to figure out which utterances mention the intended referent of DP. We compute the relevance between the hidden states $\bm{h_n}$ representing the DP and its sentence level contextual representation $\bm{cs_i}$, resulting in $rs_{n,i}$. After passing it through softmax, we obtain a sentence-level attention distribution $as_{n,i}$ as: 

\begin{equation}
rs_{n,i} = \bm{h_n}^T\bm{cs_i}
\end{equation}
\begin{equation}
as_{n,i} = \frac{e^{rs_{n,i}}}{\sum\nolimits_{i=1}^m e^{rs_{n,i}}}
\end{equation}

We then conclude sentence-level background information $\bm{s_n}$ of DP as a weighted sum of contextual utterance representations:

\begin{equation}
\bm{s_n} = \sum_{i=1}^m as_{n,i} \cdot \bm{cs_i}
\end{equation}

\subsubsection{Update DP representation}
This operation updates 
representation of the DP by combining the original DP information $\bm{h_n}$ with sentence-level background knowledge $\bm{s_n}$ through a linear transformation as: 

\begin{equation}
\bm{hs_n} = \bm{W^{2d\times4d}} [\bm{h_n}, \bm{s_n}] + \bm{b^{2d}},
\end{equation}

\noindent The updated DP state $\bm{hs_n}$ will be used in subsequent word level inference.

\subsubsection{Identify DP referents}
This operation aims to capture the word sequence that represents the intended referent of the DP.
We cast the updated DP state $\bm{hs_n}$ onto the space of contextual words $\bm{cw_{i,j}}$ by measuring the similarity $rw_{n,i,j}$ between them. The casting operation also serves as a regularization measure to restrict the inference space. The similarities are fed into the softmax function to give a word-level attention distribution $aw_{n,i,j}$  for each contextual word as:

\begin{equation}
rw_{n,i,j} = \bm{W^{1\times2d}}(\bm{hs_n} \odot \bm{cw_{i,j}})+ b^{1}
\end{equation}
\begin{equation}
aw_{n,i,j} = \frac{e^{rw_{n,i,j}}}{\sum\nolimits_{j=1}^k e^{rw_{n,i,j}}},
\end{equation}

\noindent The attention distribution $aw_{n,i,j}$ also provides an interpretation for DP recovery results based on the words it attends to.

Finally, we derive the word-level representation $\bm{w_n}$ from word states $\bm{cw_{i,j}}$. Word-level attention $aw_{n,i,j}$ is used to compute a weighted sum of contextual word states $\bm{cw_{i,j}}$ of each utterance, yielding the referent representation $\bm{tw_{n,i}}$. Then, sentence-level attention $as_i$ further filters out irrelevant referents,
yielding the final referent representation $\bm{w_n}$ as:

\begin{equation}
\bm{tw_{n,i}} = \sum_{j=1}^k aw_{n,i,j} \cdot \bm{cw_{i,j}}
\end{equation}
\begin{equation}
\bm{w_n} = \sum_{i=1}^m as_{n,i} \cdot \bm{tw_{n,i}}
\end{equation}

\subsection{Output Layer and Training Objective}
The output layer predicts the recovery result based on DP state $\bm{h_n}$ 
and referent representation $\bm{w_n}$. We feed the concatenation of these two parts into a 2-layer fully connected softmax classifier to give a categorical probability distribution over 17 candidate pronouns as:

\begin{equation}
\bm{\alpha_n} =tanh (\bm{W_1} \cdot [\bm{h_n; w_n}] + \bm{b})
\end{equation}
\begin{equation}
P(y_n|x_n,\!\bm{C})\! =softmax (\bm{W_2} \cdot \bm{\alpha_n} + \bm{b})
\end{equation}
 
We train our model by minimizing cross-entropy between the predicted label distributions and the annotated labels for all sentences. The training objective is defined as:

\begin{equation}
\resizebox{0.86\hsize}{!}{$\!loss\! = \!-\! \sum_{l \in N}\!\sum_{n=1}^s\!\delta(y_n|x_n,\!\bm{C})\! \log(P(y_n|x_n,\!\bm{C})),$}
\end{equation}

\noindent where $N$ represents all training instances, $s$ represents the number of words in each sentence; $\delta(y_n|x_n,c)$ represents the annotated label of $x_n$.

\section{Datasets and Experimental Setup}
\subsection{Datasets}
We perform experiments on three Chinese conversational datasets.
\begin{itemize}
\setlength{\itemsep}{-2pt}%
	\item Chinese text message (SMS) data, which consists of 684 SMS/Chat files. We use the same dataset split as \cite{Yang:15}, which reserves $16.7\%$ of the training set as a held-out development set to tune the hyper-parameters and evaluate the model on a separate test set.
	
	\item OntoNotes Release 5.0, which is used in the CoNLL 2012 Shared Task. We use a portion of OntoNotes 5.0 that consists of transcripts of Chinese telephone conversation (TC) speech. The 9,507-sentence subset of OntoNotes only has coreference annotations for anaphoric zero pronouns. In order to formulate the task as an end-to-end sequence labeling task, we annotate all the dropped pronouns following annotation guidelines described in \cite{Yang:15}. 
	
	\item BaiduZhidao, which is a question answering dialogue dataset used in \cite{zhang:neural}. It contains 11,160 sentences that are annotated with 10 types of concrete dropped pronouns.
\end{itemize}

Figure~\ref{corpus-statistics} shows the statistics of each type of pronouns in the three data sets.
According to \cite{Yang:15}, 5 out of 15 types of dropped pronouns in the SMS data set are \textit{abstract pronouns} that do not correspond to any actual pronoun in Chinese. These can refer to an event (Event), the previous utterance (Previous Utterance), or an generic or unspecific entity (Generic). The other two abstract pronouns are used to indicate the subject of  an existential construction (Existential) or a pleonastic subject (Pleonastic).
The other ten types of Chinese pronouns are classified as concrete pronouns. The same 15 types of pronouns are used to annotate the TC data. We can see that the BaiDuZhidao data set only has concrete pronouns.

\subsection{Evaluation metrics}
We use the same evaluation metrics as \cite{Yang:15}: precision (P), recall (R) and F-score (F).


\begin{table*}[t!]
 	\begin{center}
 		\begin{tabular}{p{4.2cm}|p{0.7cm}<{\centering} p{0.7cm}<{\centering} p{0.7cm}<{\centering}| p{0.7cm}<{\centering} p{0.7cm}<{\centering} p{0.7cm}<{\centering}|p{0.7cm}<{\centering} p{0.7cm}<{\centering} p{0.7cm}<{\centering}}
 			\hline
 			\multirow{2}{*}{Model} & \multicolumn{3}{c|}{Chinese SMS} & \multicolumn{3}{c|}{TC of OntoNotes} & \multicolumn{3}{c}{BaiduZhidao} \\ \cline{2-10} 
			 &  P(\%) &  R(\%) &  F & P(\%) & R(\%) & F & P(\%) & R(\%) & F \\ \hline
			MEPR \cite{Yang:15} & 37.27 & 45.57 & 38.76 & - & - & - & - & - & - \\ 
			NRM$^*$ \cite{zhang:neural} & 37.11 & 44.07 & 39.03 & 23.12 & 26.09 & 22.80 & 26.87 & 49.44 & 34.54 \\ 
			BiGRU & 40.18 & 45.32 & 42.67 & 25.64 & 36.82 & 30.93 & 29.35 & 42.38 & 35.83 \\ \hline
			NDPR-rand & 46.47 & 43.23 & 43.58 & 28.98 & 41.50 & 33.38 & 35.44 & 43.82 & 37.79 \\ 
			NDPR-PC-BiGRU & 46.34 & 46.21 & 46.27 & 36.69 & 40.12 & 38.33 & 38.42 & 48.01 & 41.68 \\ \hline
			NDPR-W & 46.78 & \bf 46.61 & 45.76 & 38.67 & 41.56 & 39.64 & 38.60 &\bf 50.12 &\bf 43.36 \\ 
			NDPR-S & 46.99 & 46.32 & 44.89 & 37.40 & 40.32 & 38.81 & 39.32 & 46.40 & 41.53 \\ \hline
			NDPR &\bf 49.39 & 44.89 & \bf 46.39 &\bf 39.63 &\bf 43.09 &\bf 39.77 &\bf 41.04 & 46.55 & 42.94 \\ \hline
		\end{tabular}
 	\end{center}
 	\caption{\label{comparable-result} Results in terms of  precision, recall and F-score on 16 types of pronouns produced by the baseline systems and variants of our proposed NDPR model. For NRM$^*$ \cite{zhang:neural}, we implement the proposed model as described in the paper.}
\end{table*}

\subsection{Training details}
Our model is implemented with Tensorflow.
The vocabulary is generated from the training set, which contains 17,199 word types. Out-of-vocabulary (OOV) words are represented as UNK. The BiGRU encoder uses a hidden layer of 150 units. To train the model, we use the Adam optimizer \cite{Kingma:14} with a learning rate of 0.0003. We train the model for 8 epochs and select the model with the highest F-score on the development set for testing. Dropout rate is set at 0.2 on the fully connected layers. We use uniform initialization for the weight matrices and zero initialization for biases.

\subsection{Baseline Methods and Model Variations}
We compare our proposed NDPR framework with three baselines and implement two parallel experiments:
\begin{itemize}
\setlength{\itemsep}{-2pt}%
	\item \textbf{MEPR}: This model is provided by Yang~\shortcite{Yang:15}, which uses a maximum entropy (ME) classifier with hand-crafted features to recover dropped pronouns.
	\item \textbf{NRM}: This model is proposed by Zhang~\shortcite{zhang:neural}, which uses two independent multi-layer perceptrons to locate the position of dropped pronouns and determine the type of dropped pronouns.
	\item \textbf{BiGRU}: This model encodes each word in the target sentence with a bidirectional GRU. The output representation of the BiGRU encoder is used to predict the type of the dropped pronoun. This method can be seen as a degenerate variant of our NDPR model  without the attention mechanism.
	\item \textbf{NDPR}: This model uses the BiGRU encoder with both sentence-level and word-level attention. The word embeddings are initialized with pre-trained 300-D Zhihu QA vectors \cite{Li:18} and fine-tuned when training the DP recovery model.
	\item \textbf{NDPR-rand}: Same as NDPR except that the word embeddings are randomly initialized.
	\item \textbf{NDPR-PC-BiGRU}: Same as NDPR but the encoder of utterance $X$ is replaced with PC-BiGRU.
\end{itemize}

We also perform two ablation experiments to explore the effectiveness of sentence-level attention and word-level attention:
\begin{itemize}
\setlength{\itemsep}{-2pt}%
	\item \textbf{NDPR-S}: Same as NDPR but the referent modeling layer only utilizes sentence-level attention. The output layer makes prediction with only DP state $\bm{h_n}$ and sentence level information $\bm{s_n}$.
	\item \textbf{NDPR-W}: Same as NDPR but the referent modeling layer only utilizes word-level attention. The output layer makes prediction with only DP state $\bm{h_n}$ and word-level information $\bm{w_n}$.
\end{itemize}

\section{Results and Discussion}

\begin{table}[t!]
	\begin{center}
	\small
	    \begin{tabular}{p{2.8cm}p{1.2cm}<{\centering}p{1.25cm}<{\centering}p{0.8cm}<{\centering}}
		\Xhline{1.2pt}  
		\specialrule{0em}{1.5pt}{1pt}
		 Tag & NDPR-S & NDPR-W & NDPR \\ \Xhline{0.6pt}
		\specialrule{0em}{1.5pt}{1.5pt}
        他们(masculine they) & 17.05 & \underline{23.28} &\bf 24.44 \\ 
		\specialrule{0em}{0.8pt}{0.8pt}
		她(she) & 32.35 & \underline{33.72} &\bf 35.14 \\ 
		\specialrule{0em}{0.8pt}{0.8pt}
		previous utterance & 84.90 & \underline{86.08} &\bf 87.55 \\ 
		\specialrule{0em}{0.8pt}{0.8pt}
		他(he) & 29.05 & \underline{31.20} &\bf 34.92 \\ 
		\specialrule{0em}{0.8pt}{0.8pt}
		它(it) & 25.00 & \underline{26.67} &\bf 26.95 \\ 
		\specialrule{0em}{0.8pt}{0.8pt}
		她们(feminine they) & 0 & 0 &\bf 40.00\\ \hline 
		\specialrule{0em}{1.5pt}{1pt}
		我(I) & 50.66 & \underline{50.90} &\bf 52.98 \\  
		\specialrule{0em}{0.8pt}{0.8pt}
		我们(we) & 31.49 & \underline{33.57} &\bf 34.81 \\
		\specialrule{0em}{0.8pt}{0.8pt}
		你(singular you) & 42.88 & \underline{44.15} &\bf 44.31 \\ \hline
		\specialrule{0em}{0.8pt}{0.8pt}
		pleonastic & \underline{25.89} & 22.29 &\bf 28.46 \\ 
		\specialrule{0em}{0.8pt}{0.8pt}
		generic & \underline{11.61} & 11.08 &\bf 16.83 \\ 
		\specialrule{0em}{0.8pt}{0.8pt}
		event & \underline{6.15} & 0 &\bf 16.27 \\ 
		\specialrule{0em}{0.8pt}{0.8pt}
		existential & \underline{34.17} & 30.84 &\bf 38.71 \\ \hline
		\specialrule{0em}{0.8pt}{0.8pt}
		你们(plural you) & 0 & 0 &\bf 5.41 \\ 
		\specialrule{0em}{0.8pt}{0.8pt}
		它们(inanimate they) & 16.00 &\bf \underline{19.15} & 13.89 \\ 
	    \Xhline{1.2pt} 
		\end{tabular}
	\end{center}
	\caption{\label{result-table} F-scores of our proposed model NDPR and its two variants (NDPR-S, NDPR-W) for concrete and abstract pronouns on the Chinese SMS test set.}
\end{table}

\subsection{Main Results}

Table~\ref{comparable-result} shows experimental results of three baseline systems and variants of our proposed NDPR model on Chinese SMS, TC section of OntoNotes, and BaiduZhidao. We can see that our proposed model and its variants outperform the baseline methods on all these datasets by different margins. Our best model, NDPR, outperforms MEPR by 7.63\% in terms of F-score on the Chinese SMS dataset, and outperforms NRM by 16.97\%  and 8.40\% on the OntoNotes and BaiduZhidao datasets respectively. Compared with the degenerate variant model BiGRU, our NDPR model also performs better on all three datasets, which demonstrate the effectiveness of referent modeling mechanism composed of sentence-level and word-level attention.

The experimental results also show that all components of our model have made a positive contribution, which is evidenced by the fact that the full model NDPR outperforms the other variants by a small margin. The results also show that pre-trained word embeddings have had a positive impact on the model, as shown by the higher accuracy of the NDPR model over the NDPR-rand model, where the word embeddings are randomly initialized. 

PC-BiGRU encoder seems to perform worse than vanilla BiGRU encoder, as shown by the slightly lower F-score of the NDPR-PC-BiGRU model. But it alleviates what we call the local pronoun repetition problem, the situation where a DP is recovered redundantly in a sentence before a verb and its adverbial modifier. We attribute this to that PC-BiGRU uses the context state $\bm{h_{n-1}}$ of the last word before DP instead of $\bm{h_n}$, which makes the model pay more attention to the preceding modifier.

For the BaiduZhidao dataset, the NDPR-W model actually performs better than NDPR model, as indicated  by the higher F-score for NDPR-W in the last column of Table~\ref{comparable-result}. We attribute this to the fact that there are only concrete pronouns in this dataset. The combination of ``我(I)'', ``你(singular you)'' and ``它(it)'' accounts for $94.47\%$ of the overall dropped pronoun population, for which the referent can be easily determined by word-level attention. Moreover, the fewer conversation turns in this data set mean there are few irrelevant referents that need to be filtered out by sentence-level attention.

\subsection{Ablation Study}
In this section, we dive a bit deeper and look at the impact of the attention mechanism on concrete and abstract pronouns, respectively. Table~\ref{result-table} shows the F-scores of three variants of our proposed model for each type of pronouns on the entire Chinese SMS test set. The best results among these three variants are in boldface, and  the better results between NDPR-S and NDPR-W are underlined.

\begin{figure}
	\centering
	\includegraphics[width=7.8cm, height=9.6cm]{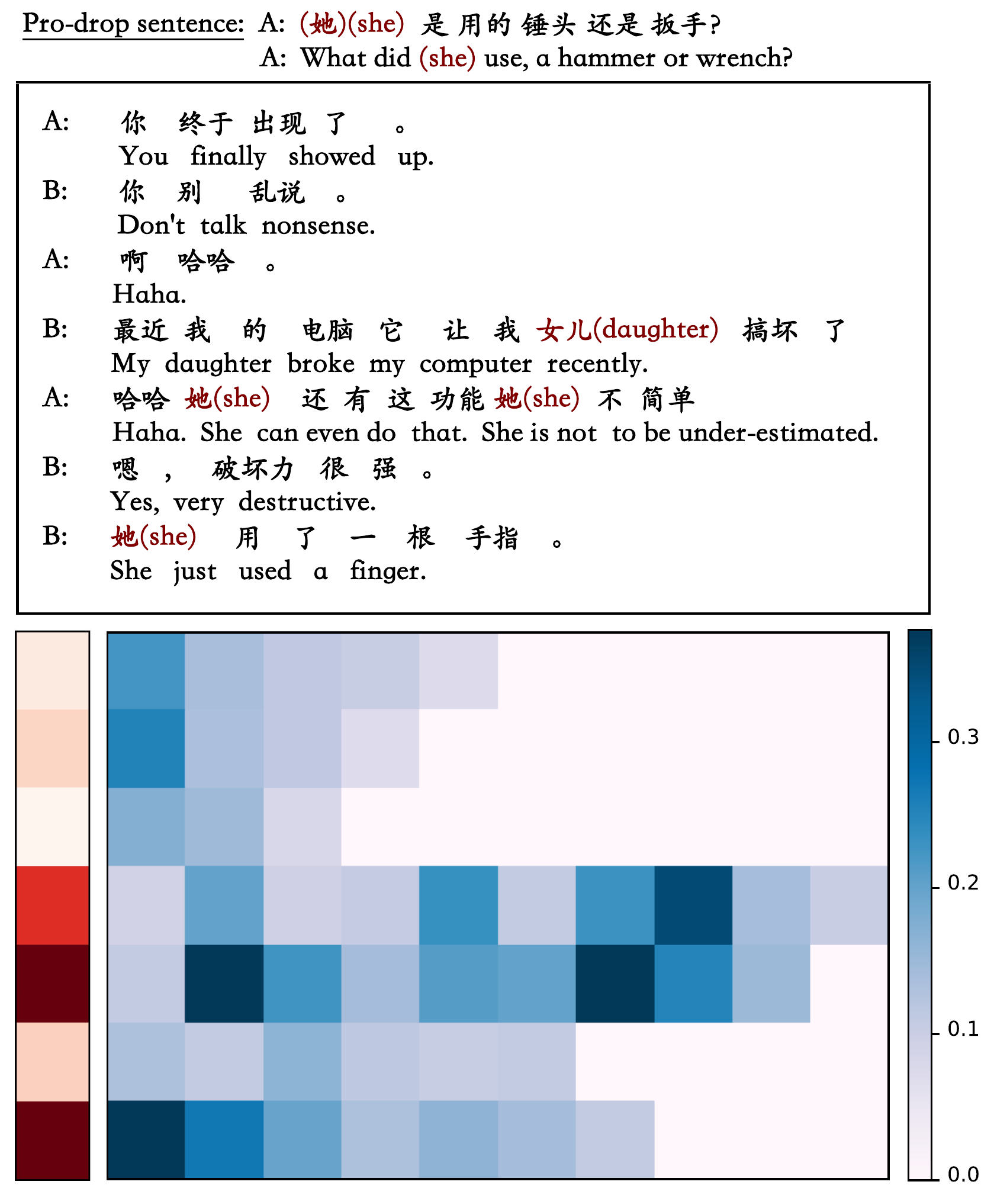}
	\caption{\label{attention} Visualization of attention in the NDPR model. The red pattern on the left shows the distribution of sentence-level attention and the blue pattern on the right shows the distribution of word-level attention. Darker color indicates higher attention weight.}
\end{figure}

We can see that for all types of concrete pronouns, the NDPR-W model outperforms the NDPR-S model by a significant margin. In general, the referent of concrete dropped pronoun is usually realized in the form of a phrase that consists of one or more word, and they can be accurately captured by word-level attention. 
The NDPR model that incorporates both word-level and sentence-level attention further improves upon NDPR-W, with the lone exception of 它们~(``they''). We believe the reason is that the sentence-level encoder cannot adequately represent the interaction between the multiple referents for plural pronouns like ~它们~ and only add noise to the representation.

However, this observation does not hold for abstract dropped pronouns. The NDPR-W performs comparably or slightly worse on four out of the five types of abstract pronouns. This is consistent with the fact that the referent of abstract pronouns like ``Event''  is often an entire sentence. 
The full NDPR model still outperforms both NDPR-W and NDPR-S for all abstract pronouns.

\begin{figure}
	\centering
	\includegraphics[width=7.8cm,height=9.8cm]{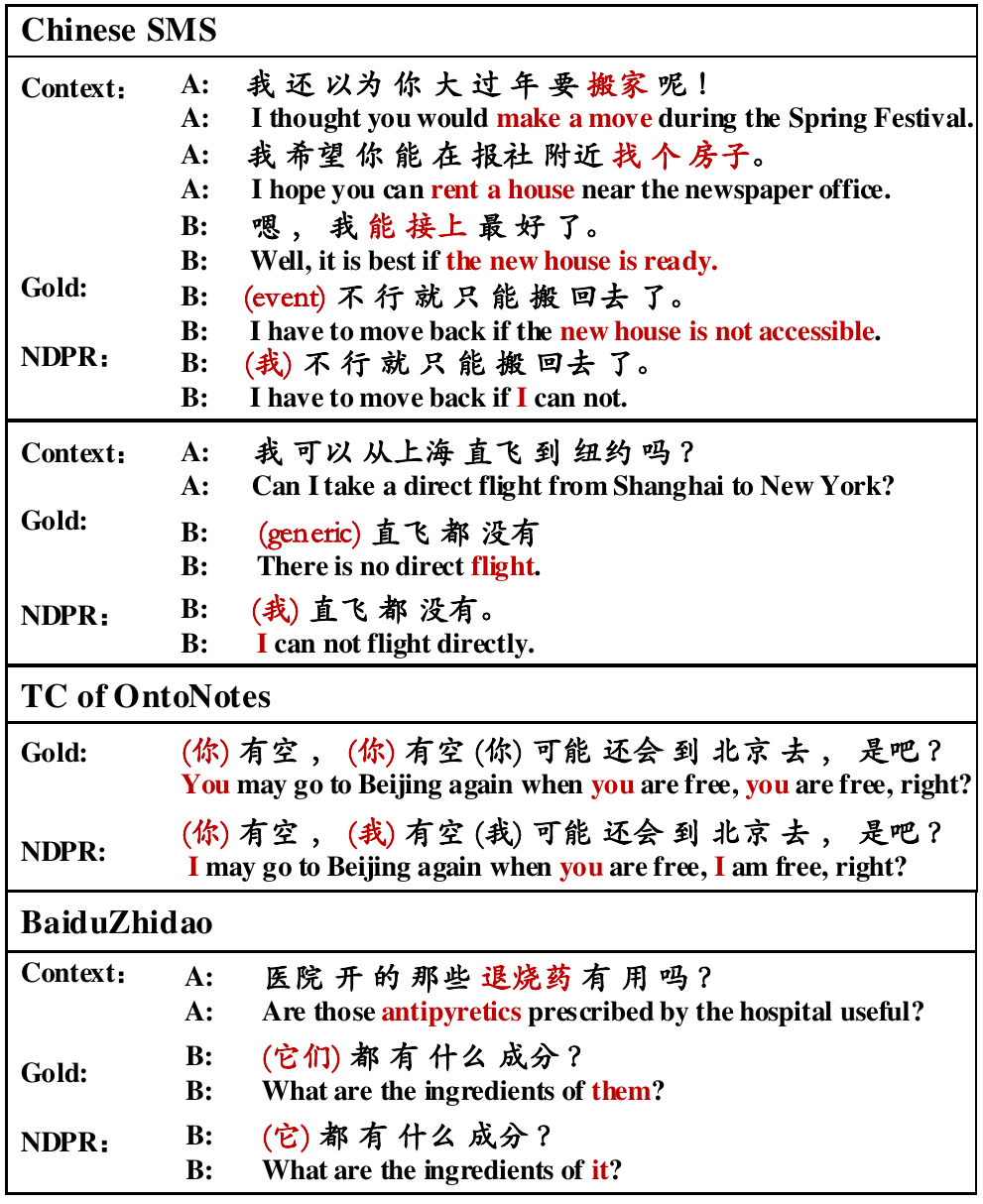}
	\caption{\label{error_case} Example errors made by NDPR on different conversational datasets.}
\end{figure}

\subsection{Attention Visualization}
Intuitively, sentence-level attention models the interaction between the dropped pronoun and each context utterance. The utterance containing the referent of the dropped pronoun should receive more attention. Similarly,  word-level attention models the interaction between each word in the context and the dropped pronoun. Thus words that describe the referent of the pronoun should receive more attention if the model works as intended. Figure~\ref{attention} shows a instance of attention distribution produced by our model.

In this example, the model correctly gives higher attention weights to three utterances that contain words that indicate the referent of the pronoun.
At word level, the model also gives higher attention weights to the specific words that indicate the referent of the dropped pronoun. For example, words such as 她~(``she'') and  女儿~(``daughter'') have received higher weights, as indicated by the darker color. This suggests that in this case the model attends to the right utterances and words and works as we have expected.

\subsection{Error Analysis}
Figure \ref{error_case} shows some typical mistakes made by our model on each genre.
For the Chinese SMS dataset, the distribution of different types of pronouns is relatively balanced (See Table~\ref{corpus-statistics}), and our model does a better job on concrete pronouns but stumbles on abstract pronouns like ``event" and ``generic" as shown in Table~\ref{result-table} as it is harder to attend to the correct parts of the context in these cases.

The TC data of OntoNotes is a transcription of telephone conversations where there are often repetitions, and our model struggles when people repeat what they said. If the same pronoun is mentioned repeatedly, our model can not capture this interaction, since each pronoun is recover independently. This suggests that one future improvement might involve using a sequence-based decoder.

In the BaiduZhidao dataset, only concrete pronouns are annotated as shown in Figure~\ref{corpus-statistics}. Pronouns like 它~(``it'') and 它们~(``they'') account for a large proportion of all dropped pronouns. 
For these two categories, the performance of our proposed method is hit-and-miss, which can be attributed to the absence of pronoun resolution common sense. 

\section{Conclusions}
We have proposed an end-to-end neural network architecture that attempts to model the interaction between the dropped pronoun and its referent in order to recover dropped pronouns in Chinese conversational data. 
Our model is based on sentence-level and word-level attention and results show that our model consistently outperforms baseline methods when evaluated on three separate datasets. We further investigate the effectiveness of different components of our model by performing ablation experiments and demonstrate the interpretability of our model using attention visualization.

\section*{Acknowledgments}
This work was supported by the National Natural Science Foundation of China (No.61702047, No.61300080) and Beijing Natural Science Foundation (No.4174098).

\bibliography{naaclhlt2019}
\bibliographystyle{acl_natbib}

\end{CJK*}
\end{document}